\documentclass[runningheads]{llncs}

\usepackage{eccv}

\usepackage{eccvabbrv}

\usepackage{graphicx}
\usepackage{wrapfig}
\usepackage{booktabs}

\usepackage[accsupp]{axessibility}  % Improves PDF readability for those with disabilities.

\usepackage{hyperref}

\usepackage{orcidlink}

\begin{document}

\title{McGrids: Monte Carlo-Driven Adaptive Grids for Iso-Surface Extraction} 

\titlerunning{McGrids}

\author{Daxuan Ren\inst{1}$^*$ \and
Hezi Shi\inst{1}$^*$ \and
Jianmin Zheng\inst{1} \and
Jianfei Cai\inst{12}}

\authorrunning{D. Ren, H. Shi, et al.}

\institute{College of Computing and Data Science, Nanyang Technological University, Singapore \and
Department of Data Science \& AI, Monash University\\
\email{\{daxuan001, hezi001\}@e.ntu.edu.sg, \\
\ asjmzheng@ntu.edu.sg, jianfei.cai@monash.edu}}

\maketitle

\def\thefootnote{*}\footnotetext{These authors contributed equally to this work}\def\thefootnote{\arabic{footnote}}

\begin{center}
    \centering
    \captionsetup{type=figure}
    \includegraphics[width=0.99\textwidth]{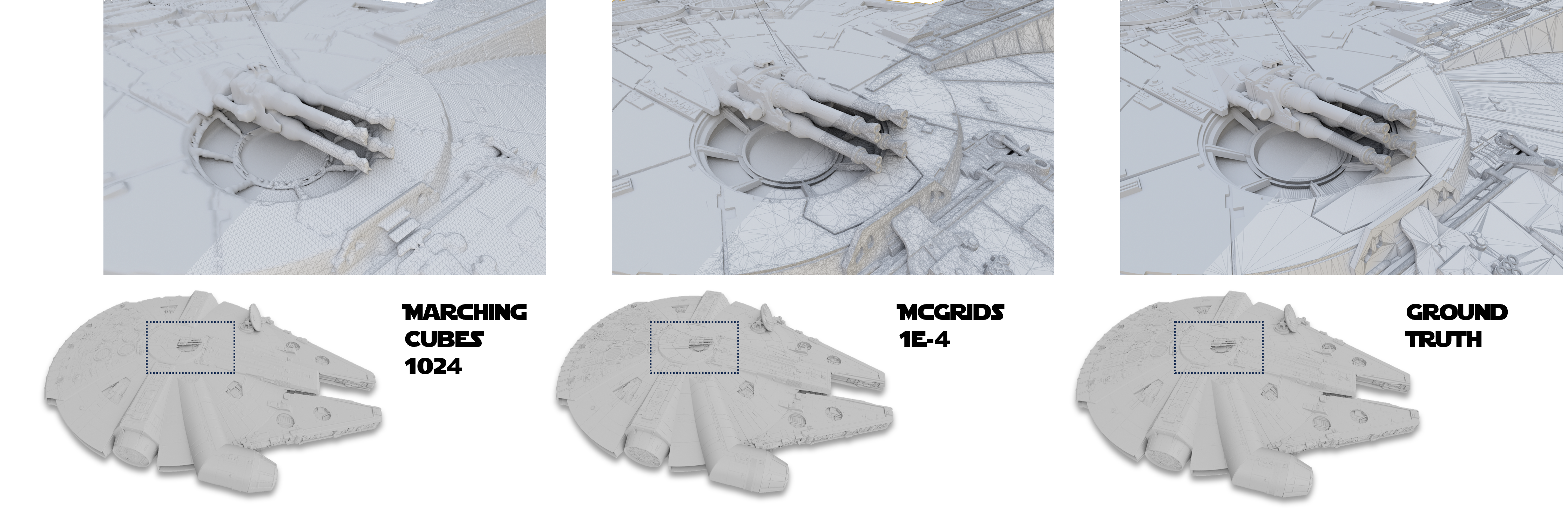}
    \captionof{figure}{McGrids can efficiently extract a highly adaptive and accurate mesh from complex implicit fields. \textbf{Left:} Mesh extracted using Marching Cubes with a voxel grid resolution of 1024 (RAM usage: $>$150GB, \#Queries: 1,073,741,824). \textbf{Middle:} Mesh extracted by McGrids (RAM: $<$ 10 GB, \#Queries: 62,012,058).  \textbf{Right:} Ground truth mesh. \textbf{Top row:} Zoom in for better visualization.}
\label{fig:teaser}
\end{center}%

\begin{abstract}
Iso-surface extraction from an implicit field is a fundamental process in various applications of computer vision and graphics. When dealing with geometric shapes with complicated geometric details, many existing algorithms suffer from high computational costs and memory usage. This paper proposes McGrids, a novel approach to improve the efficiency of iso-surface extraction. The key idea is to construct adaptive grids for iso-surface extraction rather than using a simple uniform grid as prior art does. Specifically, we formulate the problem of constructing adaptive grids as a probability sampling problem, which is then solved by Monte Carlo process. We demonstrate McGrids' capability with extensive experiments from both analytical SDFs computed from surface meshes and learned implicit fields from real multiview images. The experiment results show that our McGrids can significantly reduce the number of implicit field queries, resulting in significant memory reduction, while producing high-quality meshes with rich geometric details. 
\end{abstract}

\section{Introduction}
Implicit fields have played a pivotal role in recent advancements in computer vision and graphics such as novel view synthesis, shape reconstruction, and text-to-3D generation~\cite{mildenhall2020nerf,chen2022recovering,poole2022dreamfusion}. The flexibility of implicit fields in representing shapes with various topologies has enabled them to be seamlessly integrated into modern deep-learning approaches. However, in many engineering and graphics applications, mesh remains as the default choice. Thus, it is still necessary to extract the explicit surface mesh from the underlying implicit representations. 

Marching cubes is arguably the most popular and well-optimized method for extracting surface meshes from a grid of implicit values due to its simplicity~\cite{marchingcubes1987,marchingTets1991,Chen2021NMC,liao2018dmc}. To obtain a high-quality mesh that well preserves geometric details, the resolution of the implicit value grid must be dense. This was not an issue, as a dense grid used to be generated directly from sensor data or a fast implicit function. However, with the recent trend of representing an implicit field as a neural network, the computational time for building a dense grid becomes unacceptable since each grid point needs to undergo a complete network forward propagation to compute its value. Particularly, when the resolution in Marching Cubes is increased for extracting a highly detailed mesh, the computation time and memory usage will increase cubically. Tab.\ref{tab:query_time} shows the computation time for computing the value grids of some popular methods under different resolutions.

\setlength{\columnsep}{10pt}
\setlength{\intextsep}{4pt}
\begin{wraptable}{r}{0.5\textwidth}
\centering
\caption{Network query time in seconds for different networks at different resolutions, with a batch size of 512.} %\rdx{maybe good to also include memory usage}}
\centering
\resizebox{\linewidth}{!}{
\begin{tabular}{lllllllll}
 \hline
 Res & NGLOD~\cite{takikawa2021nglod} & NeuS~\cite{wang2021neus} & NeRF~\cite{mildenhall2020nerf}  & OccNet~\cite{sima2023occnet}\\ [0.5ex] 
 \hline
 128 & 0.30 & 1.21 & 1.44 & 2.77\\
 256 & 2.31 & 9.49 & 11.36 & 22.03\\
 512 & 18.72 & 75.77 & 90.51 & 174.80\\ 
 1024 & 152.28 & 611.74 & 721.90 & 1399.62\\ 
 2048 & 1230.11 & 1,218.45 & 5,759.82 & 11102.25\\ 
 
  \hline
\end{tabular}
}
\label{tab:query_time}
\hspace{-5pt}
\end{wraptable}

This paper aims to improve the performance of iso-surface extraction from complicated geometric shapes defined by neural implicit representations in terms of computational time and memory usage. Rather than refining the mesh extraction method, which is already highly optimized for pre-computed implicit value grids, our attention shifts towards optimizing the process of grid construction for improved performance. Our primary idea is that an efficient method should not require a dense grid and should be able to extract high-quality surface meshes with a minimum number of network queries. The ideal grid should only be dense in the regions that are near object surfaces and contain geometric details; in contrast, it should be sparse in empty or ``flat'' areas. This will ensure better accuracy in extracting iso-surfaces while minimizing the number of network queries. 

To construct the optimal grid for iso-surfacing, we cast our problem as a probability sampling problem. 
From a geometrical point of view, an Iso-Surface is a set of 3D points, where each point has an implicit value equal to a user-specified iso-value. It can also be viewed as the result of many sampling trials from a probability density function that is zero everywhere except on the object's surface. To provide a concrete example, consider function $f(x) = x$ in the 1D space. The ``0 iso-surface'' of this function is the origin. From a geometric aspect, it is the intersection point of $f$ and $y=0$. However, this ``0 iso-surface'' can also be viewed as a point sampled from a probability density function (PDF), i.e., a Dirac delta function
$\delta_\gamma(x) = \lim_{\gamma\to 0} \frac{1}{\|\gamma \|\sqrt{\pi}} e ^{-(x/\gamma)^2} $. Note that instead of just sampling points on the iso-surface, we can also use a slightly large $\gamma$ value to sample near the ``iso-surface''. In our case, for an implicit function $f(x)$ and its $\alpha$ iso-surface, we define a probability density function $\delta(x)=\frac{1}{\gamma + \lvert f(x)-\alpha \rvert} $, where $\gamma$ is a hyperparameter that controls the scattering of sampled points near the iso-surface. Apart from a dense grid near iso-surface, we also want the grid to be denser in the region with more geometry details, and sparse in ``flat'' regions. This can be achieved by using a curvature $\kappa$ weighted version of $\delta$. Thus, the overall PDF that can be used to sample points of an adaptive grid becomes: 
\begin{equation}
d(x)=\kappa \frac{1} {\gamma + \lvert f(x)-\alpha \rvert} 
\label{eq:density_function}
\end{equation}
Once $d(x) $ is defined, the generation of an adaptive grid can be cast into a sampling process that outputs the grid.  

Since $d(x)$ is still implicitly defined, we cannot directly sample from this distribution. We solve this problem by a Monte Carlo process. Our approach starts with a continuous uniform distribution and iteratively performs two steps: \textbf{1)} we sample the PDF $d(x)$ to generate a set of sample points; \textbf{2)} we refine the PDF estimation using the sampled point positions and their implicit values. Once the iteration converges, we can use all the sample points from previous iterations to construct an adaptive tetrahedron grid. Since grids are generated from Monte Carlo processes, we call them {\em McGrids}.

In summary, the main contributions of the paper are twofold:
\begin{itemize}
    \item We propose a novel approach to enhance the efficiency of performing iso-surface extraction. The key idea is to construct McGrids, adaptive grids that respect the underlying geometric shape, for iso-surfacing. McGrids significantly reduce the number of neural implicit function queries and memory usages in the process of iso-surface extraction (see Fig.~\ref{fig:teaser}). 
    \item We cast the problem of generating McGrids as a Monte Carlo sampling process, which provides an efficient and accurate solution for complicated shapes with fine geometric details, as demonstrated in experiments. The proposed approach can seamlessly serve as a plug-and-play tool for many applications that require iso-surface extraction.
\end{itemize}

\section{Related Work}
This section briefly reviews related work in neural implicit representations, iso-surface extraction, and  Monte-Carlo process.

\subsection{Neural Implicit Representations}
Neural implicit representations are a class of methods that leverage neural networks to define geometric shapes. Different from traditional voxel, point cloud, or mesh representations, neural implicit representation uses a neural network to approximate an implicit function. 
%, such as the signed distance field, which is defined over the domain.
These representations can be broadly classified into three main categories: density field~\cite{mildenhall2020nerf, Zhang2020nerfpp, Barron2021mipNerf}, distance field~\cite{park2019deepSDF, Gropp2020regular, wang2021neus, Yariv2021volsdf, takikawa2021nglod}, and occupancy field~\cite{Mescheder2019OccNet, sima2023occnet, Peng2020convOccu}.
A density field takes a query position and outputs its the volume density. The representative works such as NeRFs leverage the density field in conjunction with a color field, facilitating volume rendering.
Distance fields represent 3D shapes as level sets of deep networks, mapping 3D coordinates to a scalar value denoting the distance to the nearest surface. Occupancy fields map continuous spatial functions to constant occupancy values, simplifying the learning process for neural networks. 

These representations, being independent of spatial resolution, demonstrate a remarkable ability to capture intricate geometric details. However, despite their theoretically unlimited resolution, they still requires algorithms such as marching cubes to extract a surface mesh with finite resolution.

\subsection{Iso-surface Extraction}
The iso-surface extraction methods are generally categorized into three categories: surface tracking based methods, Delaunay refinement methods, and spatial decomposition methods. %In this section, we review the representative works in different approaches and illustrate the advantages and disadvantages of the methods.

\subsubsection{Surface tracking approaches.}
The surface tracking approaches refer to the algorithms that start with seed points and grow the mesh by attaching new triangles to the boundaries.
The pioneering work is Marching Triangles~\cite{Hilton1996MT}, which is a region-growing method to place mesh vertices according to the local surface geometry. It uses 3D Delaunay constraints to produce topologically correct meshes with well-shaped triangular faces. \cite{Schreiner2006} utilizes curvature information to construct a guidance field for triangle generation. It produces an adaptive mesh that preserves the sharp features but requires additional knowledge of the gradient of the implicit field. The surface tracking methods easily fail to extract multi-component shapes, as they are highly dependent on initialization. Meanwhile, they can hardly be parallelized due to the region-growing way.

\subsubsection{Delaunay refinement approaches.}
The mesh generation algorithms based on Delaunay refinement aim to produce well-shaped triangles with a lower bound on triangle angles. Delaunay refinement approaches start by constructing a Delaunay triangulation or Delaunay tetrahedralization, and then refine the mesh by inserting new vertices. The positions of new vertices are carefully selected to ensure adherence to the boundaries and enhance the overall mesh quality.

The pioneer work~\cite{Chew1993} introduced an algorithm for constructing restricted Delaunay triangulations based on farthest points strategy. Then, various methods have aimed for adaptive triangular faces. 
For instance, \cite{Bern1999} introduced a sampling theory based on the local feature size for smooth surfaces. 
\cite{Oudot2005} applied the farthest point strategy in $C^2$ smooth surface, and demonstrated a practical way to compute local feature size. 
\cite{Cheng2007} proposed a different algorithm that does not require any local feature size computations. Instead, it checks for violations of a topological property by ensuring a homeomorphism between input and output. 
\cite{Wang2016OnVS} focused on optimizing the regularity of extracted mesh. By inserting and adjusting vertices of extracted surface, their goal is to extract isotropic surface mesh from implicit fields.
Although Delaunay refinement approaches produce high-quality meshes suitable for various numerical methods, they are not optimal for high-resolution iso-surfacing due to their computational expense.

\subsubsection{Spatial decomposition approaches.}
The spatial decomposition approaches consider iso-surface extraction as a root-finding problem, typically employing uniform or adaptively uniform grids to access the surface. As each grid is independent and can easily be parallelized, these methods allow rapid isosurface generation. %Our method falls into this category.
In particular, the widely used Marching Cubes method~\cite{marchingcubes1987} uniformly partitions the space into a set of cubes and examines the values at eight corners of each cube to determine the existence and shape of iso-surfaces from a lookup table. The Marching Tetrahedrons method~\cite{marchingTets1991} uses tetrahedral elements and examines the values at four vertices of each tetrahedron to determine the iso-surface. It can reduce the ambiguity that arises in Marching Cubes, especially in regions with complex geometry or where iso-surfaces intersect grid cells in a nontrivial manner. 
However, these methods struggle to capture complex geometric details and waste numerous queries in void spaces. 

Dual contouring techniques~\cite{tao2002dualcontour} are more adept at capturing sharp features. They use the first-order deviation to locate points corresponding to sharp features. They generally require normal vectors and extra calculations to solve a quadratic error minimization problem.
To generalize uniform sampling to adaptive sampling, hierarchical data structures like Octree are used, where the grids intersecting the surface are refined to a finer resolution~\cite{wielson2004DMC, Kazhdan2007octree}. While these approaches save computations in void spaces, they tend to overly refine the surface, even when iso-faces could be obtained by interpolating one large grid. %Another problem is that, by the end of grid subdivision, these methods require additional implicit field queries to generate dual grids for extracting valid surface meshes.

Recently, neural marching cubes~\cite{Chen2021NMC} and neural dual contouring~\cite{chen2022MDC} were proposed, which extract iso-surfaces with data-driven approaches. These methods, which do not require normal vectors as input, leverage learned features to accurately recover sharp features of the surfaces. However, because they are trained on specific datasets, their performance may degrade when applied to new data that are significantly different from the training distribution. 
 Another data-driven method, \cite{Nissim2023voromesh}, partitions space into watertight Voronoi cells in a differentiable manner. It initializes dense uniform grids, generates Voronoi diagrams, and optimizes point locations. However, the surfaces extracted by this method are brick-like, lacking smoothness and accuracy.

\subsection{Monte Carlo methods}
%Kenann Crane
Monte Carlo methods refer to a class of statistic processes or numerical algorithms that find results through repeated random sampling. The basic approach involves generating samples randomly from a probability distribution over the domain, then performing deterministic computation on the samples, and finally aggregating the results. The underlying idea is to use sampling to simplify the complexity of the problems and use randomness to find a solution that might be deterministic in principle. Monte Carlo methods are widely used in numerical computation such as solving differential equations, solving combinatorial problems, optimization and simulation~\cite{caflisch_1998}. For example, Monte Carlo sampling was used for PDE-based geometry processing~\cite{Crane2020, Marschner2021, Nabizadeh2021, Crane2022} with great potential.

In this paper, we propose a random sampling process to estimate the density function for optimal space tessellation. Our core idea aligns with the Monte Carlo process but is different from conventional iso-surfacing approaches that follow simple uniform grid patterns for discretization.
\section{Methods}

\begin{figure*}[th]
  \centering
  \includegraphics[width=0.9\textwidth]{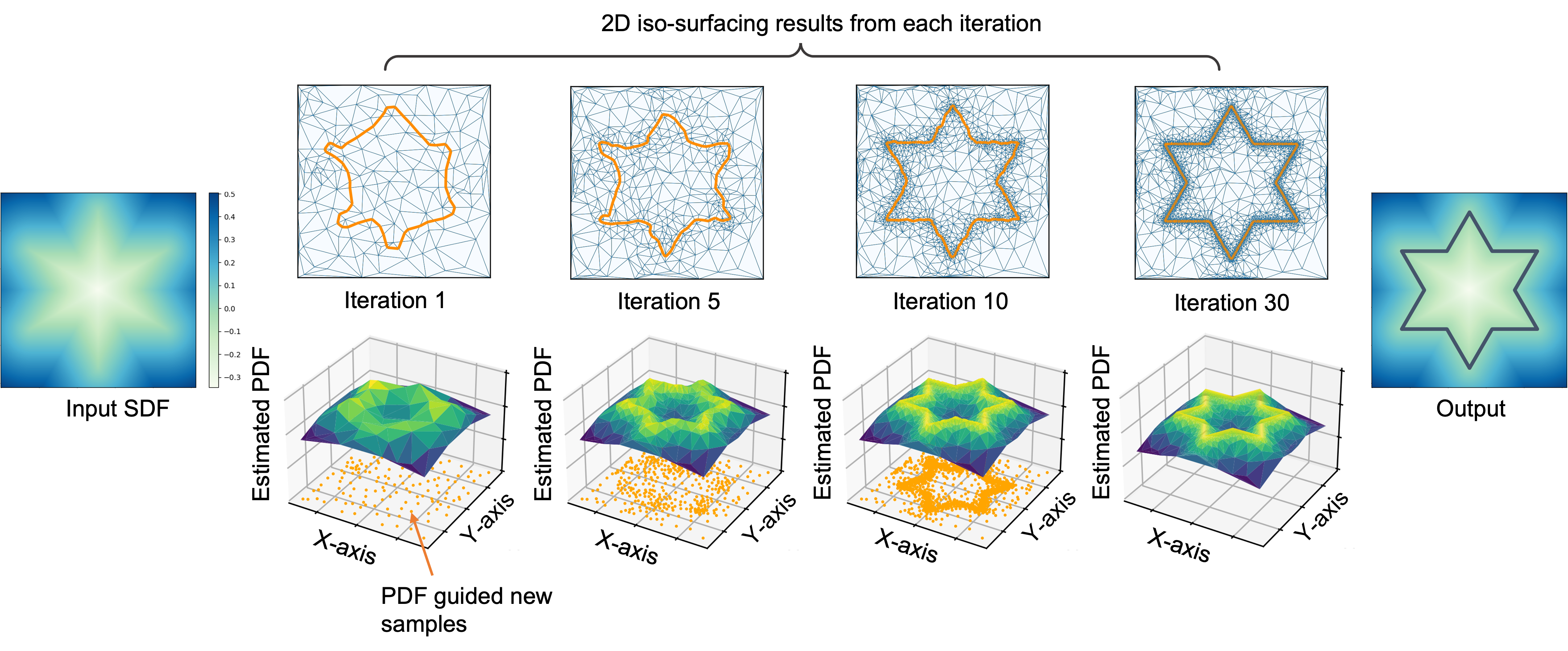}
  \caption{A simplified 2D illustration of McGrids that builds an adaptive grid around the object's iso-surface. With given implicit function (SDF), we define a probability density function (PDF), i.e. Eq.~\eqref{eq:density_function}, to guide the generation of grid points. We initialize approximation with a uniform distribution, then iteratively sample new grid points according to \eqref{eq:density_function}. The positions and the implicit values of new samples are used to refine the PDF estimation. Once the iterations converge, the tetrahedron grids with all the sample point positions are used to extract the final mesh using marching tetrahedron.} \label{fg:overview}
\end{figure*}

\subsection{Overview}
\label{sec:31}
The fundamental problem addressed in this paper is that given an implicit field $f:\mathbb{R}^3\rightarrow\mathbb{R}$, we want to extract the surface mesh that represents the $\alpha$-iso-surface. 
The mainstream approach is to extract the iso-surface from a simple uniform grid, where the positions of the mesh vertices are determined by the zero-level set of a piecewise linear function approximating the implicit field on the grid edges. While prior methods focus on constructing this piecewise linear function, we focus on creating an adaptive non-uniform grid. Theoretically, the optimal grid for iso-surface extraction is the one where the implicit function in each grid cell can be well approximated by a linear combination of the values at the grid points. This results in a piecewise linear approximation of the 3D implicit field, ensuring each triangular face accurately represents the local iso-surface while minimizing grid points to reduce computation.
%ensuring that each estimated triangular face accurately represents the local iso-surface. Meanwhile, it is desirable to minimize the number of grid points to avoid unnecessary computations. 
To this end, we propose a novel approach for iso-surface extraction with two key technical components. First, an iterative Monte-Carlo process generates adaptive grids (McGrids). Second, a refinement step adds points near the iso-surface, speeding up convergence. Finally, marching tetrahedra is used to extract the surface mesh from McGrids. Fig.~\ref{fg:overview} illustrates the proposed McGrids. Details of the three components are elaborated below.

\subsection{McGrids Generation}
Our Monte Carlo process starts with a continuous uniform distribution in a user-defined region and initializes an empty Delaunay tessellation. In each iteration, points are sampled from the distribution, and then locally relaxed to ensure grid regularity, more in Sec.~\ref{sec:cvt}. After relaxation, the points are inserted into the Delaunay tessellation. The implicit values of the points computed by querying the implicit field are also recorded. After obtaining implicit values, we can compute the probability density at each grid point using Eq.~\eqref{eq:density_function}. 
We then interpolate the point density values to the tetrahedra in the entire space and normalize them by their volumes. These tetrahedra form a piecewise linear approximation of the underlying probability density function ({PDF}). With the updated PDF, we can then sample new points from the space. This iterative process continues until reaching the user-defined number of iterations.

\noindent\textbf{Data structure}. 
We leverage Delaunay tetrahedralization to partition the grid points into non-overlapping tetrahedra, which ensures that no point lies inside the circumsphere of any tetrahedron by maximizing the minimum angles at vertices. Tessellating the space in such a way generates well-shaped tetrahedra, which benefits numerical interpolation for density values. Meanwhile, by a simple duality transform, the Delaunay tetrahedralization can be transformed to Voronoi diagram which further enables the optimization of the sample point distribution through algorithms like Centroidal Voronoi Tessellation (CVT).

\setlength{\columnsep}{10pt}
\setlength{\intextsep}{5pt}
\begin{wrapfigure}{r}{0.5\textwidth}
  \centering
  \includegraphics[width=.5\textwidth]{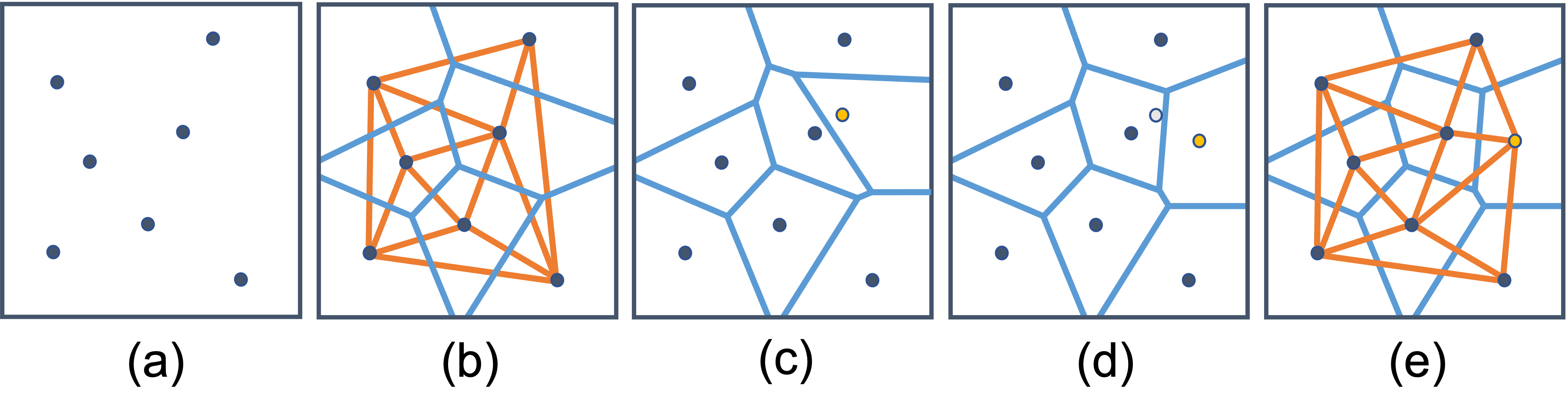}
  \caption{\small Underlying data structure of McGrids and the process of inserting a new point into McGrids. From left to right: (a) Sample points; (b) Delaunay triangulation and dual Voronoi diagram; (c) Insert a new point to McGrids; (d) CVT relaxation that moves the newly inserted point to its Voronoi centroid; (e) Updated McGrids. 
  }
  \label{fg:dual_graph}
\end{wrapfigure}

These data structures play a crucial role in guiding the generation and refinement processes of McGrids. We store all the computed point values in the Voronoi cells, as they have a one-to-one mapping to the McGrid points. The Voronoi diagram also facilitates the local relaxation process to generate more regular tetrahedra. The dual of the Voronoi diagram is the tetrahedron grid, where we interpolate the density value and sample the PDF. In our implementation, the Delaunay triangulation and the corresponding Voronoi diagram (see Fig.~\ref{fg:dual_graph}) are constructed and maintained by incrementally inserting points into the existing point set, enhancing the overall efficiency.

\noindent\textbf{PDF Estimation}. 
When new points are inserted into the Delaunay, we need to update the PDF encoded on the McGrids. The density values of the previously inserted Voronoi cells remain unchanged. However, when new points are inserted, the volume of each Voronoi cell needs to be updated. To save computation, instead of recomputing all the Voronoi cell volumes, we only update the cells that are adjacent to the newly inserted cells.

\begin{figure*}
  \centering
  \includegraphics[width=0.9\textwidth]{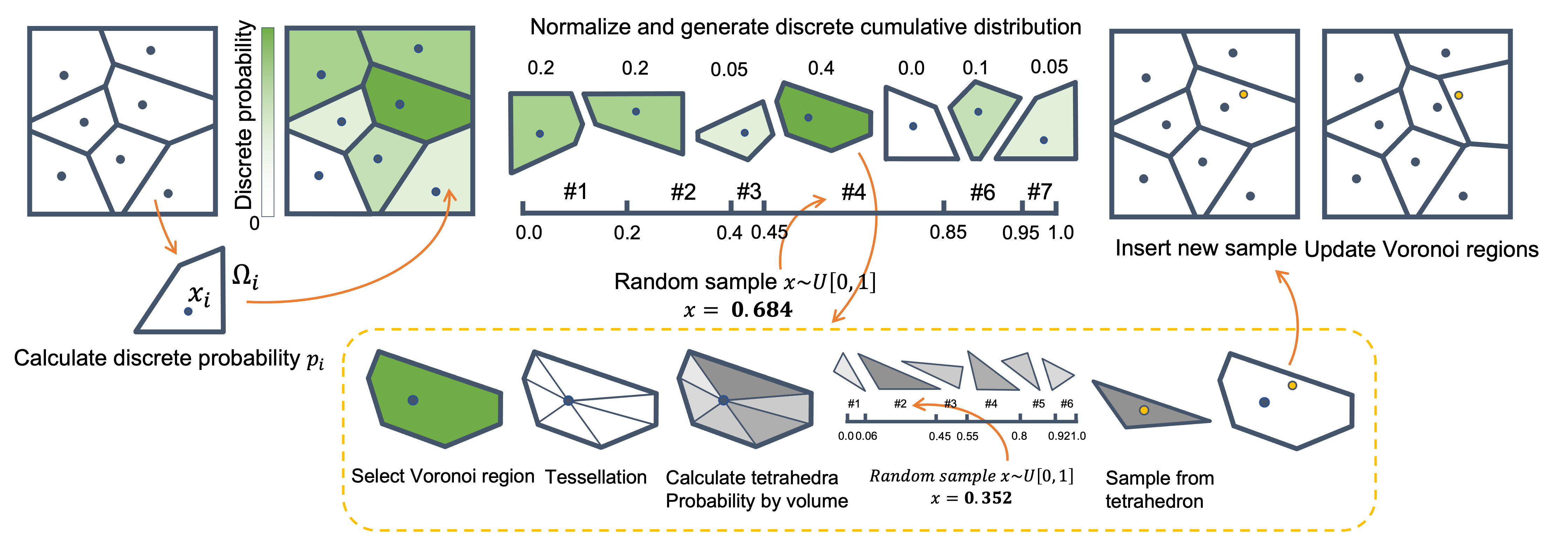}
  \caption{2D Illustration of the McGrids sampling process. For the current McGrids, we calculate discrete probabilities for each Voronoi region, and normalize and aggregate them %them Subsequently, these probabilities are normalized and aggregated 
  to form a discrete cumulative distribution. By randomly sampling a number (0.684), an indexed Voronoi region (\#4) is selected. This chosen region is then tessellated into multiple tetrahedra. After that, we calculate the volumes of these tetrahedra and form another discrete cumulative distribution for sampling. Using the selected tetrahedron (\#2), we sample a new point and insert it back into the McGrids.} \label{fg:mcsample}
\end{figure*}

\noindent\textbf{MC sampling}.
The Monte Carlo methods rely on repeated random sampling from a distribution to obtain numerical results. In our approach, we perform adaptive sampling based on the PDF defined on the previously generated McGrids iteratively. A 2D illustration is shown in Fig.~\ref{fg:mcsample}. 

In particular, with the currently constructed McGrids and the encoded PDF, we first generate a discrete probability within the given domain. For each point in the given domain, we calculate its discrete probability as
\begin{equation}
    d(x|x\in\Omega_i) = d(x_i)\cdot\frac{\int_{\Omega_{i}}x dx}{\sum^n_{i=1}\int_{\Omega_{i}}x dx} .
\end{equation}
As every point $x$ within one Voronoi region shares the same probability to be sampled, we denote this discrete probability of $\Omega_i$ by $p(i)$ and normalize it as $p(i)' = p(i)/\sum^n_{i=1}p(i)$.
After normalization, we aggregate them and create a discrete cumulative distribution ranging from $0$ to $1$. We then randomly sample a number $x \sim U[0,1]$, pick the bin in which $x$ falls, and finally select the corresponding Voronoi region to sample a point. 
To conduct a weighted random sample within a Voronoi region $\Omega_i$,  the Voronoi region is first tessellated into tetrahedra. Then, similar to the Voronoi region selection process, a tetrahedron is selected using the tetrahedra's probability density and a uniformly generated number, and then a sample point is obtained by sampling the tetrahedron.

\noindent\textbf{CVT Relaxation}. %\label{sec:cvt}
In numerical computation involving interpolation on discrete grids, most methods prefer the use of regular tetrahedra and avoid degenerated tetrahedra~\cite{Ciarlet2002}. 
This preference is rooted in the understanding that during the approximation process, it is the length of the longest edge of the tetrahedron that affects the interpolation error~\cite{Field2000}. 
%The Finite Element Method for Elliptic Problems. 1978.
%ualitative Measures for Initial Meshes. 2000.
%
To enhance the regularity of newly sampled points in the Monte Carlo sampling process, we utilize Centroidal Voronoi Tessellation (CVT) optimization~\cite{du1999cvt, du2006lloyd}. CVT optimization can be seamlessly integrated into our McGrids generation process, as it represents a unique type of Voronoi tessellations in which each site corresponds to the centroid of the Voronoi region. The goal of CVT is to achieve a more balanced distribution of points, thereby reducing the interpolation error for each tetrahedron.
In the experiment section, we will illustrate the effectiveness of CVT in accelerating convergence and improving output mesh quality.

\subsection{McGrids Refinement}
Although the distribution of the generated PDF already bears some resemblance to the implicit surface and the generated McGrids adaptively interpolate the surface, the purpose of this refinement step is to further reduce the approximation error by sampling more points to achieve lower approximation error. 
Specifically, this is accomplished by: 
\begin{itemize}
    \item ``Trimming'' the tetrahedra that do not intersect the surface (setting the probability density values of incident vertices to 0);
    \item Refining the tetrahedra that touch the surface by inserting new points inside.
\end{itemize}
Meanwhile, we refrain from refining tetrahedra that already exhibit accurate numerical results. Here we assess the quality of a tetrahedron by evaluating its midpoint, which serves as an indicator of the rough error within the tetrahedron.

\label{sec:33}
\noindent\textbf{Mid point insertion}. 
%In Eq.~\eqref{eq:density_function}, a curvature factor is used to encourage more samples at high-curvature region. 
Directly using curvature values to determine if extra points should be inserted into the tetrahedron is challenging and expensive, especially with neural implicit fields. Instead, we propose a midpoint-based approach to approximate a curvature-like value, which is effective in practice.
For a tetrahedron intersecting the surface, where four corner vertices have different signs, we perform linear interpolation to find points on edges with the desired iso-value. We then calculate the ''midpoint'' of the tetrahedron by averaging these interpolated points' positions. Since this midpoint should be on the surface if the tetrahedron is accurate and exhibits low curvature, we check the difference between its true implicit value and the iso-value. If this difference exceeds a user-defined threshold, we insert this midpoint into the McGrids.
Subsequently, we apply CVT to relax the newly inserted points, ensuring that the overall McGrids still exhibit both regularity and adaptiveness.

\noindent\textbf{Termination condition}. 
Theoretically the approximation error can be estimated based on the curvature of the field and the tetrahedral grid~\cite{Shewchuk2002}. 
In practice, we compute the approximation error in each grid by comparing the iso-value of the interpolated midpoint with its true implicit value. If the difference is smaller than a user-specified threshold, we consider the tetrahedron to be adequately refined, assuming it exhibits a smooth and linear variation within the tetrahedron. The refinement process ends when all tetrahedral grids meet the criteria. This termination condition is used in our experiments for its simplicity and efficiency, and we found it adequate in various scenarios.
In some extreme situations, e.g. fractal surfaces, we introduce additional hard constraints on tetrahedron volume. This is to prevent tetrahedra from being excessively refined, maintaining a balance between accuracy and computational efficiency.

\section{Experiments}

\begin{figure}[bh]
  \centering
  \includegraphics[width=0.92\textwidth]{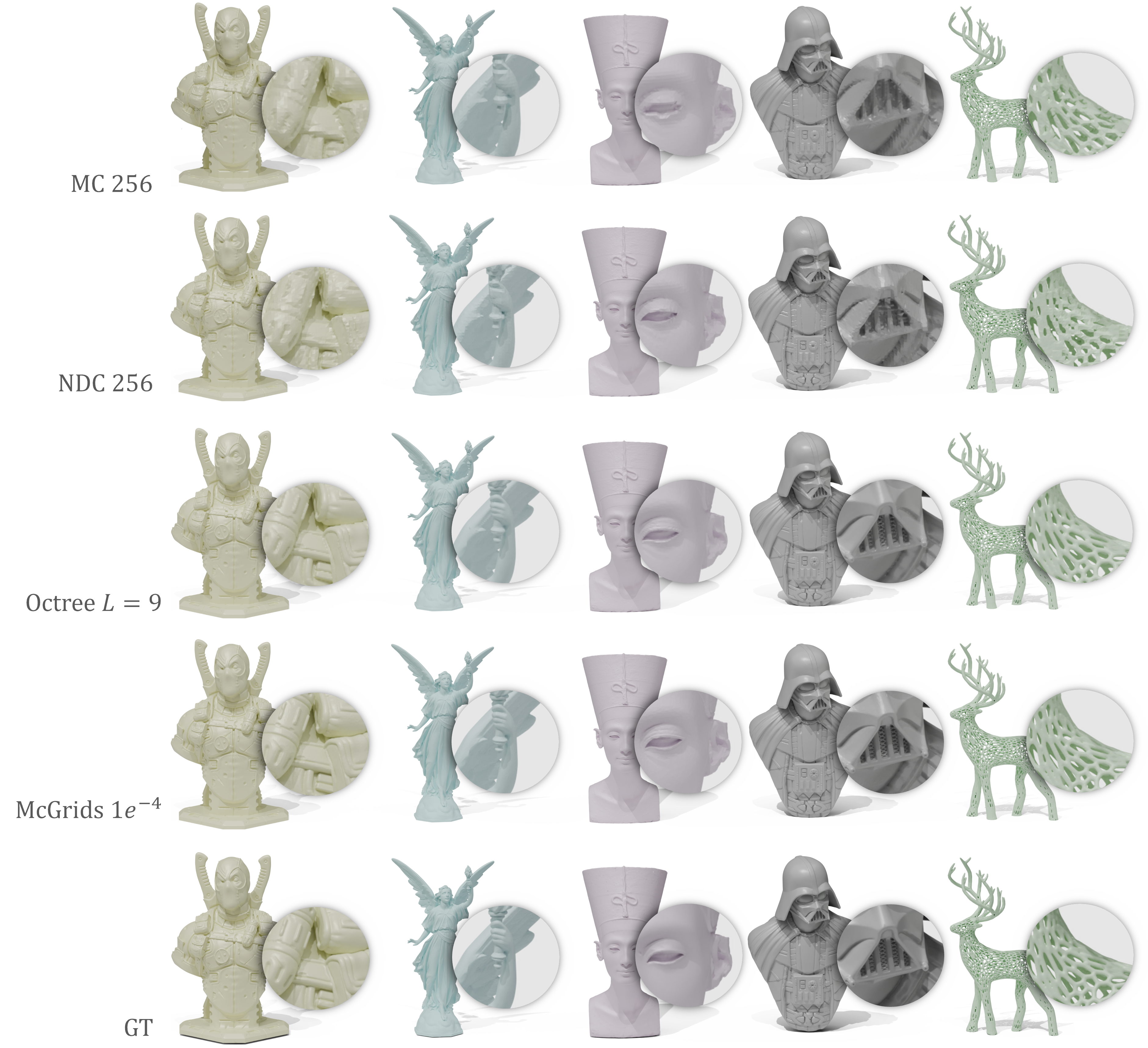}
  \caption{Visual comparisons of the extracted meshes. We compare McGrids with Marching Cubes~\cite{marchingcubes1987}, Octree method~\cite{Kazhdan2007octree}, and Neural Dual Contouring~\cite{chen2022MDC} with a resolution of $256^3$. For a fair comparison, we run McGrids with a terminating threshold of $1e^{-4}$. We can see that McGrids extracts more detailed and accurate meshes from complex implicit fields.} \label{fg:iso-surface extraction}
\end{figure}

In this section, we evaluate the accuracy and efficiency of McGrids for extracting iso-surfaces from complex shapes and neural implicit functions. We showcase qualitative visualization results as well as quantitative ones and analyze the contributions from different components in McGrids. Additionally, we explore an application of McGrids in a differentiable multiview reconstruction pipeline.
%we demonstrate that McGrids is plug-and-play and can be easily integrated into existing differentiable pipelines for improved performance.
%
\begin{table*}
\begin{center}
\caption{Quantitative iso-surface extraction results of different methods. \textbf{Top}: McGrids generally reconstructs more accurate surface meshes with much less computation time and memory consumption. \textbf{Bottom}: By varying the terminating threshold, McGrids is scalable when more detailed surface mesh is needed.}
\label{tab:comparison}

% todo: add Octree numbers
\resizebox{0.99\linewidth}{!}{
\addtolength{\tabcolsep}{6pt}  
\begin{tabular}{lllllllll}
\hline
& CD$\cdot 10^{5}\downarrow$     & NC $\uparrow$                  & ECD $\cdot 10^{3}\downarrow$   & F1$\uparrow$                   & EF1$\uparrow$                  & \#Query$\downarrow$             & Time$\downarrow$                & Memory$\downarrow$              \\ \hline
MC$_{256}$~\cite{marchingcubes1987}          & 1.27                           & 0.91                           & 0.60                           & 0.76                           & 0.50                           & 16.78M                          & 12.85s                          & 2.57G                           \\
MT$_{256}$~\cite{marchingTets1991}          & 0.61                           & 0.93                           & 2.56                           & 0.95                           & 0.41                           & 2.16M                           & 5.75s                           & 1.6G                            \\
NDC$_{256}$~\cite{chen2022MDC}         & 0.56                           & \textbf{0.94} & \textbf{0.23} & \textbf{0.96} & \textbf{0.75} & 16.78M                          & 76.89s                          & 21.76G                          \\
Octree$_{L=9}$~\cite{Kazhdan2007octree}         & 0.56                           & 0.93 & 0.37 & 0.96 & 0.71 & 1.13M                          & 2.58s                          & \textbf{0.89G}                        \\
McGrids$_{1e-3}$        & \textbf{0.54} & \textbf{0.94} & 0.47                           & \textbf{0.96} & 0.64                           & \textbf{0.44M} & \textbf{2.12s} & 0.91G \\ \hline
MC$_{512}$~\cite{marchingcubes1987}          & 0.73                           & 0.95                           & 0.24                           & 0.94                           & 0.72                           & 134.22M                         & 107.51s                         & 15.2G                           \\
McGrids$_{5\cdot 1e-4}$ & \textbf{0.51}                           & 0.95                           & 0.31                           & \textbf{0.97}                           & 0.77                           & \textbf{0.7M}                            & \textbf{4.06s}                           & \textbf{0.93G}                           \\
McGrids$_{1e-4}$        & 0.55                           & 0.96                           & 0.21                           & 0.96                           & 0.81                           & 4.3M                            & 15.86s                          & 1.26G                           \\
McGrids$_{5\cdot 1e-5}$ & 0.55                           & \textbf{0.97}                           & \textbf{0.12}                           & 0.96                           & \textbf{0.82}                           & 5.5M                            & 20.05s                          & 1.72G                           \\ \hline
\end{tabular}
}
\end{center}

\end{table*}
%

%\subsection{Evaluation of iso-surface extraction}

\subsection{Iso-surface extraction from complex shapes}
%We compare McGrids with existing iso-surface extraction methods. We will demonstrate that McGrids can extract detailed and accurate meshes with significantly fewer implicit function queries, resulting in significant memory usage reduction and faster processing time. 

\noindent\textbf{Dataset}. 
We compile a dataset consisting of complex meshes from multiple sources, including Stanford scanning repo \cite{curless1996volumetric}, Thingiverse\cite{Thingiverse.com},  NIE\cite{mehta2022level}. If source meshes are not watertight, we utilize Blender \cite{mullen2011mastering} to convert them into manifold meshes. Open3D~\cite{zhou2018open3d} is employed to query SDF values during runtime. 

%\textbf{Implementation Details}.
%McGrids is implemented in C/C++. We leverage GeoGram \cite{levy2015geogram} and CGAL\cite{fabri2009cgal} for Delaunay-related computations. McGrids is multithreaded for max performance with Intel TBB \cite{pheatt2008intel}. For easy usage and integration of McGrids with existing neural implicit representations, we also export Python APIs via PyBind11 \cite{pybind11}. Note that all the following benchmarks are conducted through the Python APIs. 

\noindent\textbf{Baselines}.
We compare our McGrids with existing iso-surface extraction methods including Marching Cubes \cite{marchingcubes1987}, Marching Tetrahedra \cite{marchingTets1991}, Octree method \cite{Kazhdan2007octree}, and Neural Dual Contouring \cite{chen2022MDC}. To ensure a fair comparison, we exclude the methods that also require normal information, \eg, \cite{tao2002dualcontour} \cite{shen2023flexicubes}.

\noindent\textbf{Evaluation Metrics}.
We evaluate McGrids from accuracy and efficiency aspects. We adopt the comparison metrics from  \cite{chen2022MDC} for measuring iso-surface extraction quality, which include $L_2$ Chamfer Distance (CD), Edge chamfer Distance (ECD), Normal Consistency (NC), F1 and Edge-F1 (EF1). To measure runtime efficiency, we measure program runtime, memory usage (system RAM and VRAM), and the number of implicit function queries. 

\begin{wrapfigure}{r}{0.3\textwidth}
  \centering
  \includegraphics[width=0.3\textwidth]{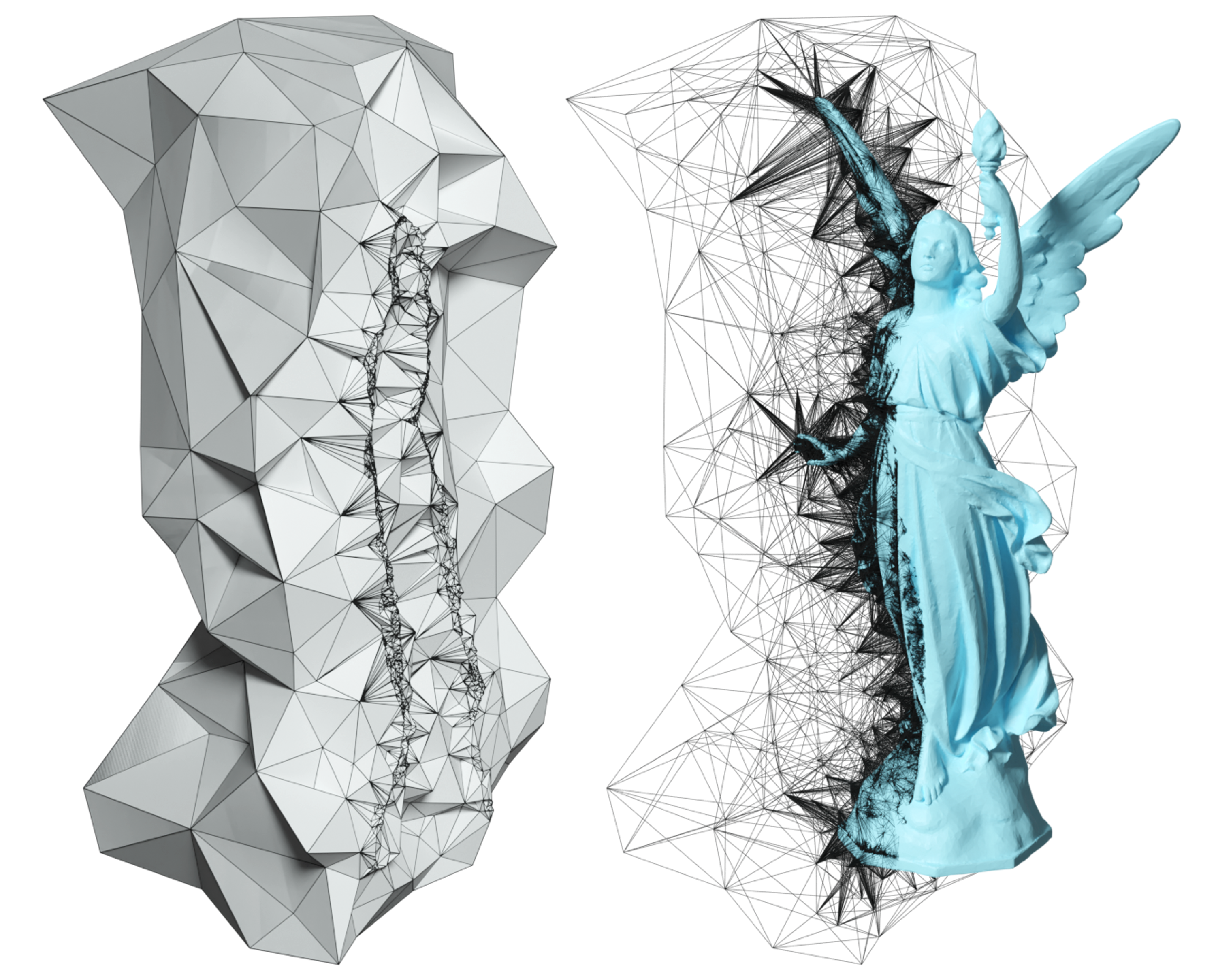}
  %\caption{ Cross-section of McGrids.}
  \label{fig:cross}
\end{wrapfigure}
\noindent\textbf{Results}. 
Tab.~\ref{tab:comparison} shows the quantitative results of different methods on the constructed dataset. We can see that our McGrids in general performs the best in most of the mesh quality metrics but with much less computation time and memory consumption. By lowering the terminating threshold, McGrids can reconstruct more detailed surface meshes with the cost of increasing runtime and memory usage.  

Fig.~\ref{fg:iso-surface extraction} gives the visual comparisons of the extracted meshes by McGrids and the two baselines. We can see that McGrids extract much more visually pleasing and detailed meshes. 
From the mesh wireframes, it can be observed that McGrids indeed extracts adaptive meshes where larger triangles are used in less detailed areas while more triangles are allocated to regions with more geometric details. In right figure, we show a cross-section view of McGrids, noticing that the tetrahedra near iso-surface and geometrically detailed areas are smaller in volume, while large at empty regions.

\subsection{Iso-surface extraction from Neural Implicit functions}
\begin{figure}
  \centering
  \includegraphics[width=0.65\textwidth]{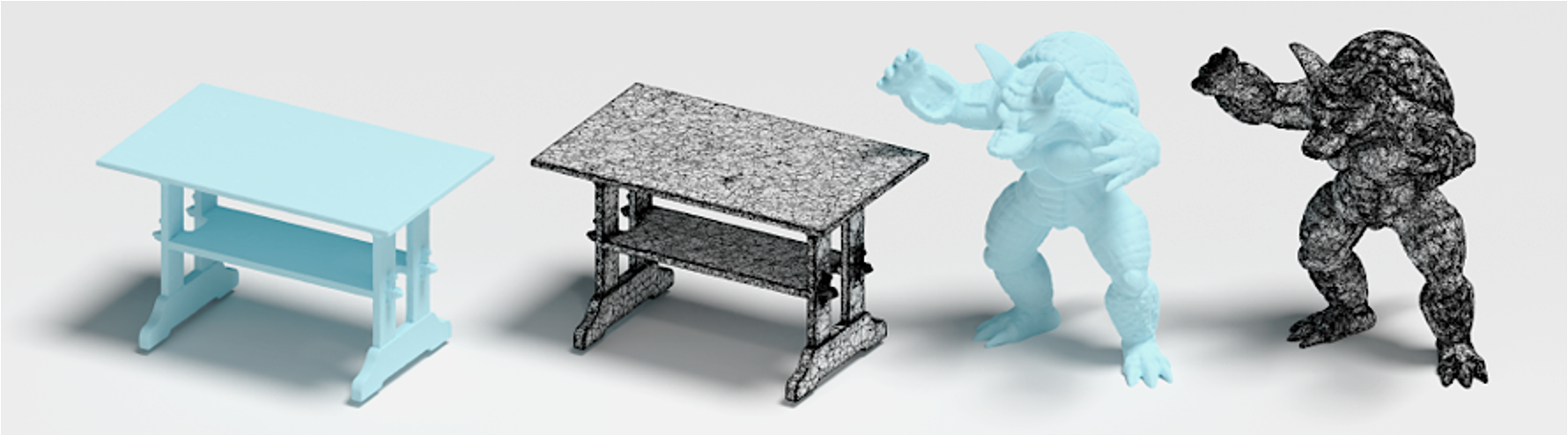}
  \caption{McGrids are capable for extracting accurate yet adaptive iso-surfaces from neural implicit fields of NGLOD \cite{takikawa2021nglod}.}  \label{fg:visual_nglod}
\end{figure}

We further evaluate McGrids for iso-surface extraction from neural implicit functions. We run McGrids on NGLOD's \cite{takikawa2021nglod} implicit fields as NGLOD is well known for its ability to encode complex geometric details. As no ground truth iso-surface exists, we show visual results in Fig.~\ref{fg:visual_nglod}, where we can see that McGrids is capable of extracting detailed geometries. Note that the extracted meshes are also adaptive, \eg, larger triangles on the tabletop and denser triangles on the edges. 

\begin{figure*}
  \centering
  \includegraphics[width=0.65\textwidth]{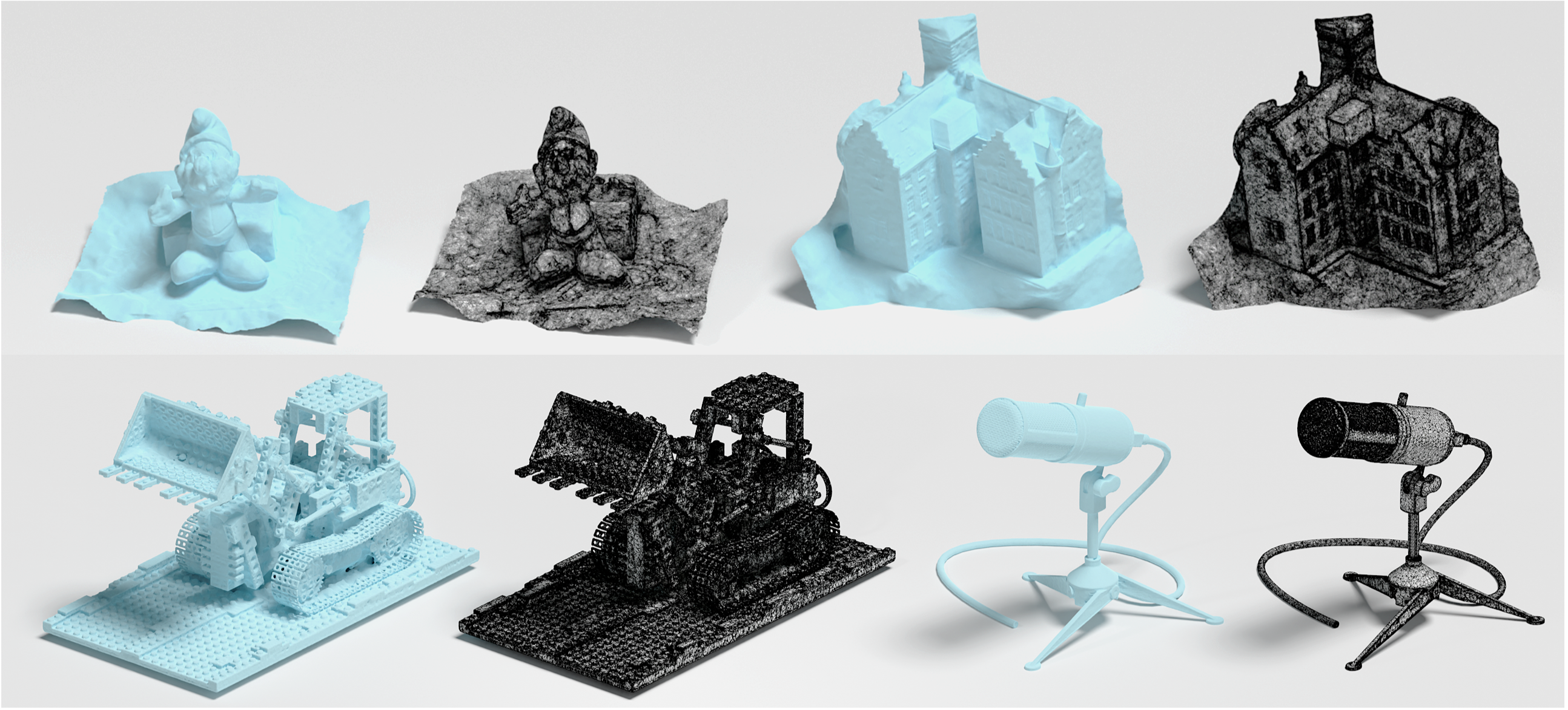}
      \caption{Visual results of extracting accurate yet adaptive iso-surfaces from neural implicit fields of NeuS~\cite{wang2021neus} (top) and VolSDF~\cite{Yariv2021volsdf} (bottom).}  \label{fg:visual_neus}
\end{figure*}

%We show more examples for iso-surface extraction from various neural implicit functions methods in 
Fig.~\ref{fg:visual_neus} gives more examples of extracting accurate yet adaptive iso-surfaces from neural implicit fields of other popular methods including NeuS~\cite{wang2021neus} and VolSDF~\cite{Yariv2021volsdf}. We render both the shaded meshes and the wireframes to visually demonstrate the adaptiveness of the extracted meshes. 

\subsection{Ablation study}
\begin{figure*}
  \centering
  \includegraphics[width=0.9\textwidth]{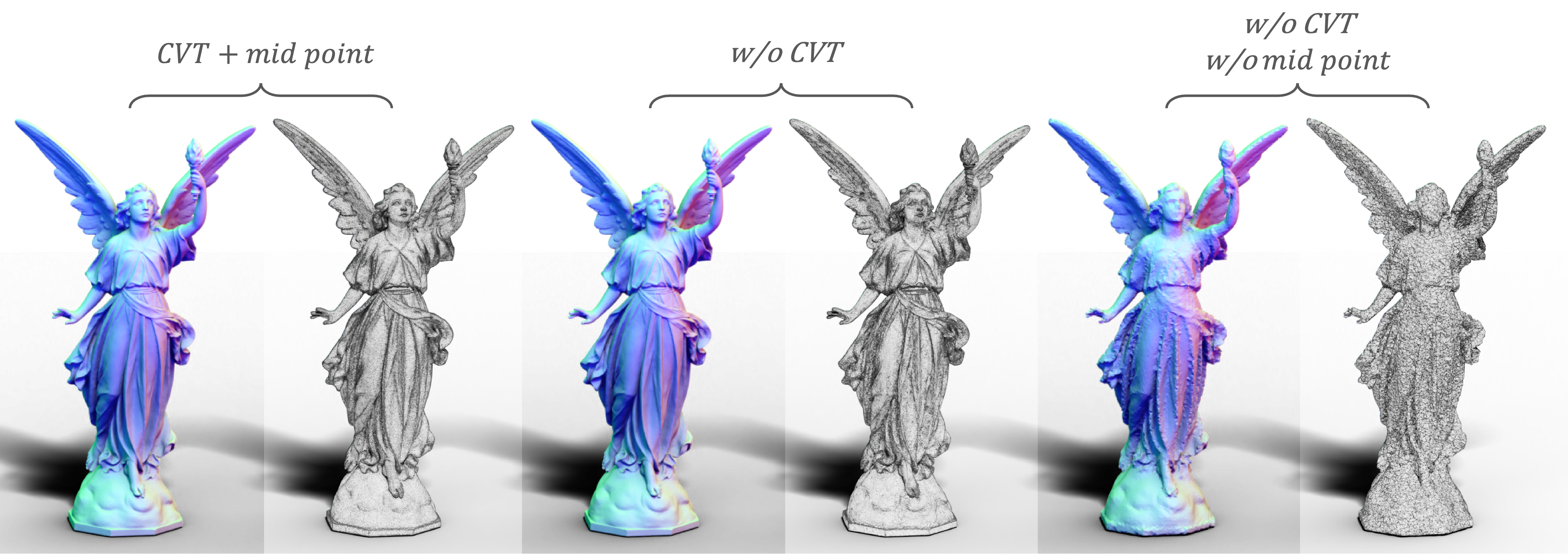}
  \caption{Analysis of CVT and mid point insertion. CVT relaxation leads to a more uniform and smooth distribution of triangular faces. Mid point insertion improves accuracy especially on regions with complex geometric details and sharp features.}
  %\caption{Comparison of the reconstruction results of McGrids with and without CVT optimization. Left two: with CVT. Right two: without CVT} 
  \label{fg:cvt}.
\end{figure*}
\noindent\textbf{CVT Relaxation}.
\label{sec:cvt}
To enhance the regularity of the generated McGrids and consequently reduce approximation errors within each tetrahedron, we employ CVT relaxation after inserting new points into the McGrids. CVT approaches are designed to optimize point distribution by minimizing the energy function of the Voronoi tessellation. There are two main categories of CVT approaches: stable~\cite{du1999cvt} and optimal~\cite{liu2009cvt}, corresponding to achieving local or global minimization, respectively. 

Our experiments show that the optimal CVT often achieves lower energy after optimization but with unacceptable computational time.
%
%the increased computational time is unacceptable. 
In contrast, stable CVT proves to be significantly faster, as well as flexible and robust, with a monotonically decreasing energy. 
Thus, we employed the classic stable CVT, Lloyd's method~\cite{du2006lloyd}. In each CVT iteration, the new site position is determined as the centroid of the corresponding Voronoi region.
Here, we run experiments on ``Lucy'' model and present qualitative comparisons in Fig.~\ref{fg:cvt}, and quantitative comparisons in Tab.~\ref{tab:cvt} for our McGrids with and without CVT.
\begin{table}
\centering
\caption{Quantitative comparisons of McGrids with and without CVT. CVT notably speeds up convergence and improves mesh accuracy, particularly for low resolution.}
\label{tab:cvt}
\resizebox{\linewidth}{!}{
\addtolength{\tabcolsep}{8pt} 
\begin{tabular}{llllllll}
\hline
& CD$(10^{5})\downarrow$ & NC  $\uparrow$ & ECD $(10^{3})\downarrow$ & F1 $\uparrow$ & EF1 $\uparrow$ & \# iter$\downarrow$ & \# query$\downarrow$ \\ \hline
$5e^{-3}$(w CVT)   & 1.84                   & 0.82           & 0.53                     & 0.74          & 0.16           & 20                               & 17,058                            \\
$5e^{-3}$(w/o CVT) & 5.51                   & 0.81           & 0.61                     & 0.74          & 0.17           & 12                               & 19,323                            \\ \hline
$1e^{-3}$(w CVT)   & 0.45                   & 0.95           & 0.11                     & 0.98          & 0.70           & 15                               & 228,472                           \\
$1e^{-3}$(w/o CVT) & 0.45                   & 0.94           & 0.12                     & 0.98          & 0.69           & 58                               & 319,185                           \\ \hline
$1e^{-4}$(w CVT)   & 0.43                   & 0.97           & 0.04                     & 0.98          & 0.85           & 18                               & 2,595,029                         \\
$1e^{-4}$(w/o CVT) & 0.43                   & 0.97           & 0.04                     & 0.98          & 0.84           & 41                               & 2,998,294                         \\ \hline
\end{tabular}
}
\end{table}

\begin{wrapfigure}{r}{0.45\textwidth}
  \centering
  \includegraphics[width=0.4\textwidth]{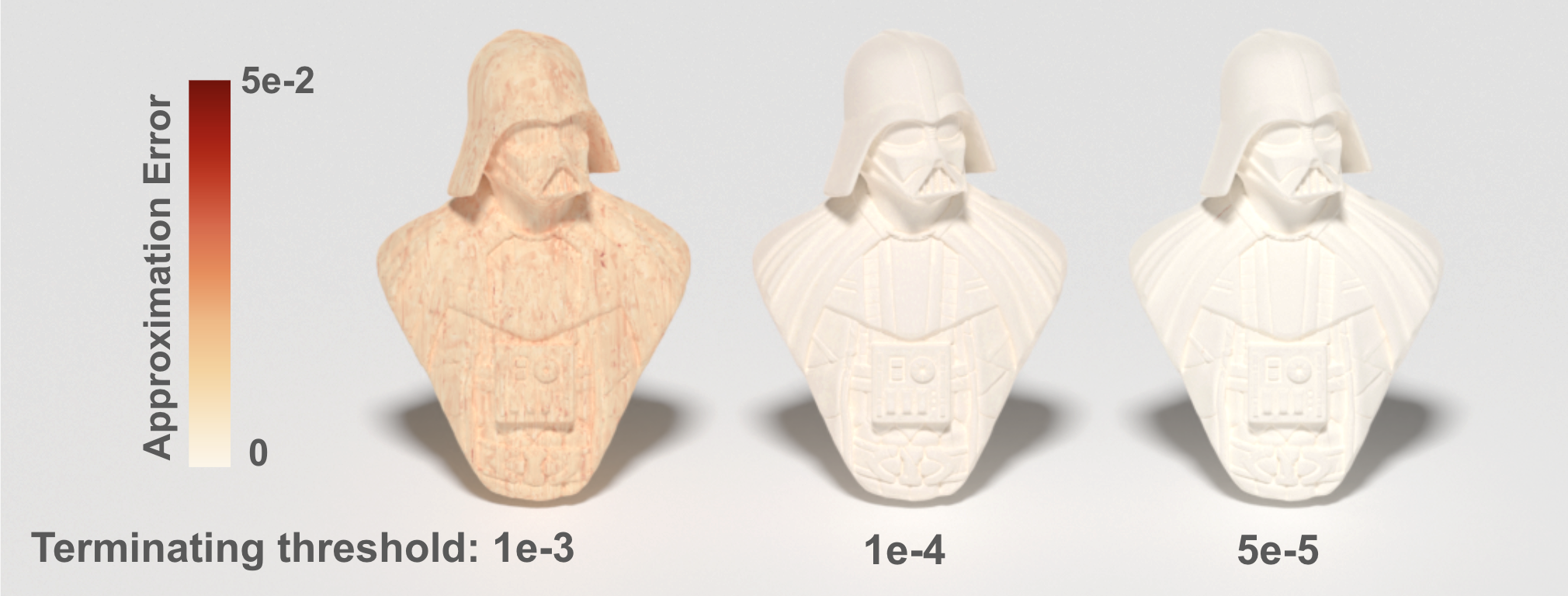}
  \caption{Visualization of errors w.r.t terminating thresholds.}
\label{fg:visual_heatmap}
\end{wrapfigure}

\noindent\textbf{Mid point insertion}.
To further enhance approximations precision, we propose a simple yet effective midpoint insertion strategy to densify the grids near the surface. Each midpoint is determined by averaging the grid-surface intersection points. After inserting the midpoint, the adjacent grids undergo a retetrahedralization process. In Fig.~\ref{fg:cvt}, a visual comparison is presented between the surfaces extracted with and without the incorporation of the midpoint insertion strategy. It is noteworthy that both sets of McGrids have an identical number of vertices. Evidently, the implementation of the midpoint insertion strategy plays a crucial role in enhancing the overall accuracy of the approximation.

\noindent\textbf{Termination condition}. 
When the approximation error of mid point of a grid is lower that the terminating threshold, or the volume of the grid is smaller that a constraint, we stop further inserting point into this grid. The Fig.\ref{fg:visual_heatmap} visualizes approximation errors with different terminating thresholds, from left to right, the heat maps reveal a systematical decrease of approximation error across the entire surface as the threshold decreases. 

%\subsection{Differentiable Multiview Reconstruction}
\subsection{Application}

\begin{wraptable}{l}{0.5\textwidth}
\centering
\caption{Multiview reconstruction results measured by $L_2$ Chamfer Distance (CD x$10^5$).}
\resizebox{\linewidth}{!}{
\begin{tabular}{llllll}
\hline
& CD$(10^{5})\downarrow$         & NC  $\uparrow$                 & ECD $(10^{3})\downarrow$      & F1 $\uparrow$                  & EF1 $\uparrow$                 \\ \hline
DMT        & 2.51                           & 0.91                           & 6.7                           & 0.51                           & 0.02                           \\
FlexiCubes & 2.23                           & 0.92                           & 8.13                          & 0.52                           & 0.03                           \\
McGrids    & \textbf{2.14} & \textbf{0.96} & \textbf{6.3} & \textbf{0.59} & \textbf{0.07} \\ \hline
\end{tabular}
}
\label{tab:multiview}
\end{wraptable}

\begin{wrapfigure}{r}{0.5\textwidth}
  \centering
  \includegraphics[width=.5\textwidth]{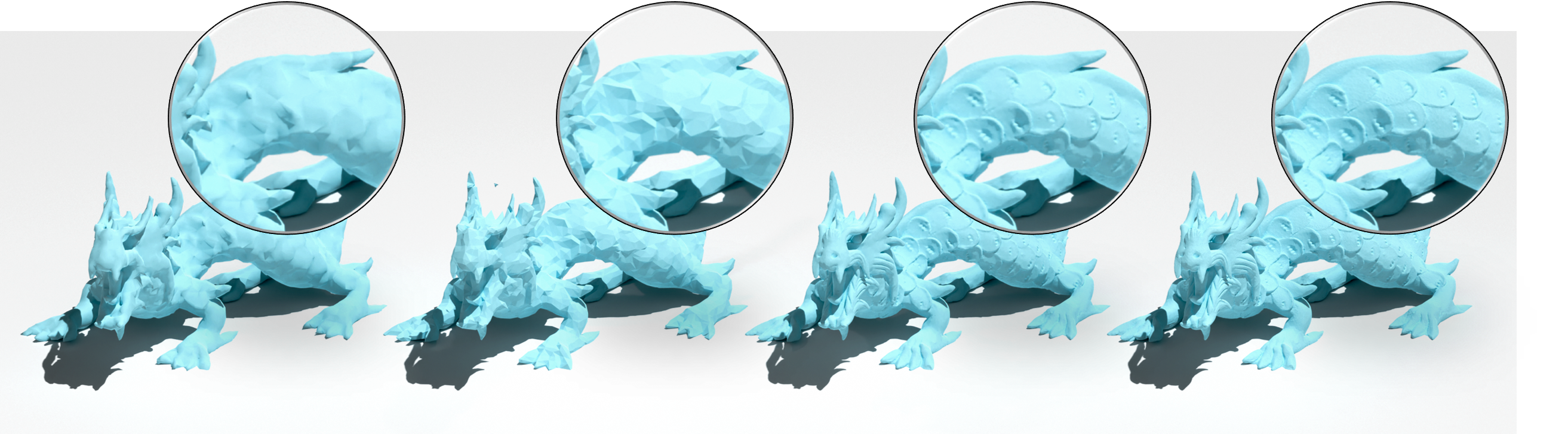}
  \caption{Visual comparisons of the reconstruction results. From left to right: DMT, FlexiCubes, McGrids, and GT.} \label{fg:multiview}
\end{wrapfigure}

Since our mesh extraction component is built on standard Marching Tetrahedron, McGrids can be integrated into a differentiable multiview reconstruction pipeline to produce accurate, adaptive yet differentiable surface mesh. Here, we replaced the mesh extraction components in DMT~\cite{shen2021DeepMT} and  FlexiCubes~\cite{shen2023flexicubes} with McGrids. Tab.~\ref{tab:multiview} shows that McGrids can achieve much more accurate reconstruction results compared to DMT and Flexicubes. Fig.~\ref{fg:multiview} illurstrates that the mesh reconstructed using McGrids is more accurate and detailed. 
Note that compared to standard DMT and FlexiCubes, which reuse fixed grids,While McGrids have a slower runtime than standard DMT and FlexiCubes due to dynamic adaptive grid computation for each iteration, they hold significant potential for applications requiring highly detailed surface meshes.

\setlength{\columnsep}{10pt}
\setlength{\intextsep}{5pt}

\section{Conclusion and Limitation}
We introduced McGrids, an adaptive tetrahedron grid generated by an iterative Monte Carlo process. It is designed to efficiently and accurately extract iso-surfaces from implicit fields. We conducted extensive experiments to demonstrate the capability of McGrids in fast extracting adaptive meshes. In addition, we integrated McGrids into a differentiable pipeline to showcase its versatility.

\noindent\textbf{Limitation.} McGrids is designed to effectively sample more grid points near object surfaces and geometric details by using our carefully designed data structure and sampling algorithm. However, this comes at the cost of additional overheads, i.e., the grid construction, compared to regular grids. Therefore, for cases that only require low-resolution meshes or situations where implicit field queries are extremely cheap, conventional methods like Marching Cubes may be a better option. In addition, since McGrids does not assume normal information, it may require a denser grid than Dual Contouring to extract sharp features, where Dual Contouring can use QEF to place vertices directly on sharp features.

\clearpage
\noindent\textbf{Acknowledgements} This work is supported by MOE AcRF Tier 1 Grant of Singapore (RG12/22), and also by the RIE2025 Industry Alignment Fund – Industry Collaboration Projects (IAF-ICP) (Award I2301E0026), administered by A*STAR, as well as supported by Alibaba Group and NTU Singapore. Daxuan Ren was also partially supported by Autodesk Singapore.
\bibliographystyle{splncs04}
\bibliography{main}
\end{document}